\documentclass[letterpaper, 10 pt, conference]{ieeeconf}

\usepackage{geometry}
\IEEEoverridecommandlockouts                                                                                

\ifdefined\DRAFT
\usepackage[draft]{graphicx} 	% for pdf, bitmapped graphics files
\else
\usepackage{graphicx} 			% for pdf, bitmapped graphics files
\fi

\usepackage{amsmath} % assumes amsmath package installed
\usepackage{amssymb}  % assumes amsmath package installed
\usepackage{algorithm}
\usepackage{algpseudocode}
\usepackage{wrapfig}
\usepackage{siunitx}
\usepackage{float}
\usepackage{subfig} 
\usepackage{balance}
\usepackage{bm}
\usepackage{epstopdf}
\usepackage{color}
\usepackage{multirow}

\newcommand{\tabref}[1]{Table~\ref{#1}}
\newcommand{\figref}[1]{Figure~\ref{#1}}
\renewcommand{\eqref}[1]{Equation~(\ref{#1})}

\newcommand{\seref}[1]{Section~\ref{#1}}

\title{\LARGE \bf
Parameter Optimization for Loop Closure Detection in Closed Environments
}

\author{Nils Rottmann$^{1}$, Ralf Bruder$^{1}$, Honghu Xue$^{1}$, Achim Schweikard$^{1}$, Elmar Rueckert$^{1}$% <-this % stops a space
 \thanks{$^{1}$Institute for Robotics and Cognitive Systems, 
	University of Luebeck,
        Ratzeburger Allee 160, 23562 Luebeck, Germany
         {\tt\small \{rottmann, bruder, schweikard, rueckert\}@rob.uni-luebeck.de}}%
}

\begin{document}

\maketitle
\thispagestyle{empty}
\pagestyle{empty}

%%%%%%%%%%%%%%%%%%%%%%%%%%%%%%%%%%%%%%%%%%%%%%%%%%%%%%%%%%%%%%%%%%%%%%%%%%%%%%%%
\begin{abstract}

\noindent Tuning parameters is crucial for the performance of localization and mapping algorithms. In general, the tuning of the parameters requires expert knowledge and is sensitive to information about the structure of the environment. In order to design truly autonomous systems the robot has to learn the parameters automatically. Therefore, we propose a parameter optimization approach for loop closure detection in closed environments which requires neither any prior information, e.g. robot model parameters, nor expert knowledge. It relies on several path traversals along the boundary line of the closed environment. We demonstrate the performance of our method in challenging real world scenarios with limited sensing capabilities. These scenarios are exemplary for a wide range of practical applications including lawn mowers and household robots.
\end{abstract}

%%%%%%%%%%%%%%%%%%%%%%%%%%%%%%%%%%%%%%%%%%%%%%%%%%%%%%%%%%%%%%%%%%%%%%%%%%%%%%%%
\section{Introduction}

\noindent Algorithms for simultaneous localization and mapping (SLAM) \cite{durrant2006simultaneous,bailey2006simultaneous} such as FastSLAM \cite{montemerlo2002fastslam}, GMapping \cite{grisetti2007improved}, \cite{grisettiyz2005improving} or RTabMap \cite{labbe2014online}, \cite{labbe2019rtab} require the tuning of a large number of parameters. A correct setting of these parameters is crucial for the performance of these algorithms \cite{abdelrasoul2016quantitative}. In general, finding convenient parameters for a certain mapping task requires prior knowledge on the structure of the environment and the robot itself. However, truly autonomous systems are expected to be able to adapt themselves to any environment and thus, being able to learn the required parameters autonomously. A well-known method for such meta-parameter learning problems is classical Reinforcement Learning (RL) \cite{sutton1998introduction}, more specifically Bayesian Optimization (BO) \cite{shahriari2015taking}, \cite{snoek2012practical}. BO is a black box optimizer that only requires a definition of a cost function. A proper definition of the cost function is critical for the success of the parameter learning procedure. For mapping algorithms, a natural choice would be to define the cost as the difference between the estimated map and the respective ground truth. However, the ground truth is not known a priori such that other cost measures have to be developed for the meta-parameter learning.\\

\noindent An area of increasing importance in the last decade is the field of low-cost robotics \cite{ciupe2014new}, \cite{hagele2016robots}. Robots such as lawn mowers or vacuum cleaners are used ubiquitously in households and work exclusively in closed environments, e.g. on a lawn or in an apartment. In general, these robots have only limited sensing capabilities due to the low-cost design. Algorithms dealing with the mapping problem for this type of robots are proposed in \cite{ozisik2016simultaneous} and \cite{choi2008line}, where sonars or infrared sensors are used and linear features required. An indoor mapping approach using a wall following scheme has been presented in \cite{zhang2010real}, where map rectification has been used under the assumption of straight wall segments.\\

\noindent Where there is an active research for SLAM approaches for autonomous vacuum cleaner, e.g. vision SLAM \cite{jung2004structured,lee2012self,lee2012vision}, autonomous lawn mowers still move randomly within the area of operation. Thereby, they use a boundary wire enclosing the working area which emits an electromagnetic signal that can be detected by the robot. Towards efficient localization and planning, a first step can be taken by mapping the enclosure. In \cite{einecke2018boundary}, a map generation approach based the loop closure detected by returning to the home station has been introduced. Thereby, the lawn mower was driving along the boundary wire while measuring movements with the wheel odometry. However, using only a single loop closure requires to distribute the error along all estimated positions equally. Hence, detecting additional loop closures is favorable for a robust mapping approach. In \cite{rottmann2019loop}, the authors proposed a loop closure detection approach for low-cost robots based on odometry data only. The data is collected when the robot is following the boundary of the closed environment. The performance of this approach depends highly on the correct meta-parameter setting which requires a priori knowledge about the closed environment. Hence, to enable truly autonomous behavior the robot has to learn the parameter by itself such that it can adapt to any arbitrary closed environment. Therefore, we developed a RL approach for learning meta-parameters under the assumption that the average distance traveled by the robot along a closed environment is equal to its circumference. We demonstrate the performance and robustness of our approach in different challenging simulation and real world scenarios.\\

\noindent The contributions of the paper are three-fold. First, we adapt and improve the method introduced in \cite{rottmann2019loop} by introducing relative error measurements for each loop closure using the \textit{Iterative Closest Point} (ICP) approach \cite{besl1992method}. Second, we insert a feasibility check in order to cope with recurrent symmetric structures and third, we introduce a RL scheme for learning the meta-parameters to enable true autonomous behavior. For our approach, we require that the robot is able to travel several times along the boundary line of the closed environment, e.g. by using a perimeter wire. \\

\noindent We start by summarizing and adapting the mapping method from \cite{rottmann2019loop}, \seref{se:Background}. In \seref{se:ParameterLearning}, we derive the RL procedure for meta-parameter learning. The procedure is divided into two stages, parameter learning for loop closure detection and pose graph optimization. We evaluate our approach in simulations and on real data in \seref{se:Results} and in \seref{se:Conclusion} we conclude.

\section{Mapping Procedure}\label{se:Background}

\noindent As the robot follows the boundary line of the closed environment, e.g. by means of the electromagnetic wire signal, a path based on the robot odometry data can be recorded. This path can then be transferred into the well-known pose graph representation \cite{grisetti2010tutorial}. Loop closures can be identified by comparing the neighborhoods of the pose graph vertices with each other, e.g. due to shape comparison. Based on the identified loop closing constraints the pose graph can be optimized by reducing the sum of weighted residual errors. Both, finding good loop closing constraints and optimizing the pose graph are strongly dependent on the correct parameter tuning. In the following, we shortly recapture pose graph representation as well as the general idea for detecting loop closing constraints. In \tabref{tab:variables}, we listed our notations for the different variables used throughout this paper.

%\begin{figure}[tb]
%\centering
%\includegraphics[width=0.44\textwidth]{pics/TraceSegmentation.eps}
%\caption{Path Segmentation: In the left panel the original odometry data with 37687 data points is shown. In the right panel the path segmentation with 137 dominant points, marked as red dots, is presented.}
%\label{fig:ExampleTraceSegmentation}
%\end{figure}

\begin{table}[b]
\centering
\caption{Variable definitions used throughout this paper.}
\label{tab:variables}
\begin{tabular}{lll}
$\boldsymbol{p}$ & $\mathbb{R}^{3}$ & poses\\
$\boldsymbol{x}$ & $\mathbb{R}^{2}$ & positions in meters\\
$\varphi$ & $\mathbb{R}$ & orientations in rad\\
$\boldsymbol{R}$ & $\mathbb{R}^{2 \times 2}$ & two-dimensional rotation matrix \\
$\boldsymbol{\xi}$ & $\mathbb{R}^{3}$ & relative measurements \\
$\boldsymbol{P}$ & $\mathbb{R}^{3 \times 3}$ & cov. matrix to the noise of the rel. measurements \\
$N$ & $\mathbb{N}$ & number of odometric constraints \\
$M$ & $\mathbb{N}$ & number of loop closing constraints \\
$L_{\text{NH}}$ & $\mathbb{R}$ & neighborhood length in meters\\
$c_{\text{min}}$ & $\mathbb{R}$ & minimum comparison error\\
$\gamma_1,\gamma_2$ & $\mathbb{R}$ & pose graph optimization parameters\\
$U$ & $\mathbb{R}$ & circumference of the closed environment in meters \\
$u$ & $\mathbb{R}$ & path distance between loop closing pairs in meters \\
$\Delta \varphi$ & $\mathbb{R}$ & difference in orientation in rad \\
$\varphi_{\text{cycle}}$ & $\mathbb{R}$ & feasibility check parameter \\
\end{tabular}
\end{table}

\subsection{Pose Graph Representation}\label{se:PGORepresentation}
\begin{figure}[bt]
\centering
\includegraphics[width=0.48\textwidth]{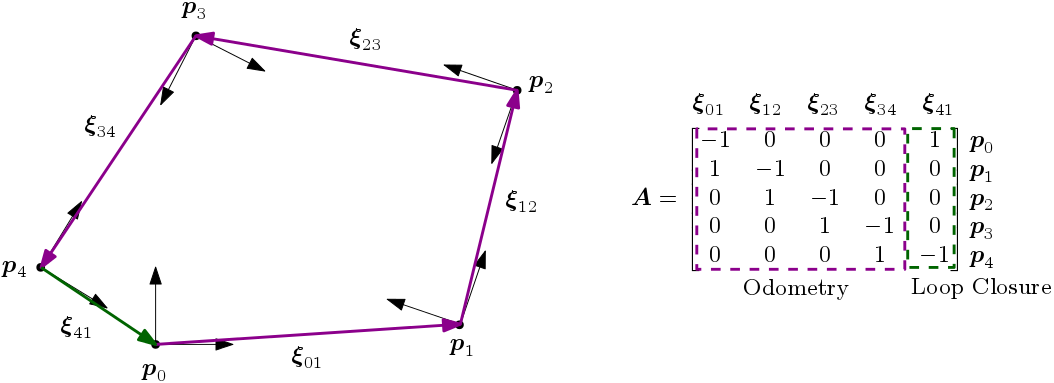}
\caption{Pose graph with five vertices connected with five edges. Four of the edges are odometric constraints and one is a loop closing constraint. On the right, the incidence matrix is shown divided into the parts containing the odometric constraints and the loop closing constraints.}
\label{fig:IncidenceMatrix}
\end{figure}

\noindent Let ${\boldsymbol{p} = \{\boldsymbol{p}_0,\dots,\boldsymbol{p}_N\}}$ be a set of $N+1$ poses representing the position and orientation of a mobile robot in a two dimensional space, hence $\boldsymbol{p}_i = [\boldsymbol{x}_i^{\top}, \varphi_i]^{\top}$. Here, $\boldsymbol{x}_i \in \mathbb{R}^2$ is the cartesian position of the robot and $\varphi_i \in [-\pi,\pi]$ the corresponding orientation as an euler angle with the integer $i = 0\mathit{:}N$. The relative measurement between two poses $i$ and $j$ is then given as
\begin{equation}
\label{eq:RelativeMeasurement}
\boldsymbol{\xi}_{ij} = \begin{bmatrix}
\boldsymbol{R}_i^{\top} (\boldsymbol{x}_j - \boldsymbol{x}_i) \\
\varphi_j - \varphi_i
\end{bmatrix} = \boldsymbol{p}_j  \circleddash \boldsymbol{p}_i,
\end{equation}
where $\boldsymbol{R}_i = \boldsymbol{R}_i(\varphi_i)$ is a planar rotation matrix and $\circleddash$ the pose compounding operator introduced by \cite{lu1997globally}. The pose graph is then a directed graph ${\mathcal{G}(\mathcal{V},\mathcal{E})}$ with $N+1$ vertices, representing the poses, and $N + M$ edges, representing the relative pose measurements. In our case, these pose measurements are composed of $N$ odometric constraints and $M$ loop closing constraints. In \figref{fig:IncidenceMatrix}, an example of a pose graph with four odometric and one loop closing constraint is shown. The connection between the vertices by the edges can be compactly written using an incident matrix $\boldsymbol{A}$, which is exemplarily shown on the right in \figref{fig:IncidenceMatrix}.\\

\noindent To account for noise in the relative pose measurements, we include zero mean Gaussian noise ${\boldsymbol{\epsilon}_{ij} \sim \mathcal{N}(\boldsymbol{0}, \boldsymbol{P}_{ij})}$, where
\begin{equation}
\label{eq:RelativeMeasurementsNoisy}
\hat{\boldsymbol{\xi}}_{ij} = \boldsymbol{\xi}_{ij} + \boldsymbol{\epsilon}_{ij} ,
\end{equation}
denotes the with noise corrupted relative pose measurements. The overall optimization problem is then to minimize the sum of weighted residual errors ${\boldsymbol{r}_{ij}(\boldsymbol{p})}$ with respect to the pose estimates $\boldsymbol{p}$,
\begin{equation}
\label{eq:OptimizationProblemGeneral}
\min_{\boldsymbol{p}} \sum_{(i,j) \in \mathcal{E}} ||\boldsymbol{r}_{ij}(\boldsymbol{p})||^2_{\boldsymbol{P}_{ij}},
\end{equation}
where
\begin{equation}
\label{eq:ResidualError}
||\boldsymbol{r}_{ij}(\boldsymbol{p})||^2_{\boldsymbol{P}_{ij}} = [(\boldsymbol{p}_j  \circleddash \boldsymbol{p}_i) - \boldsymbol{\hat{\xi}}_{ij}]^{\top} \boldsymbol{P}_{ij}^{-1} [(\boldsymbol{p}_j  \circleddash \boldsymbol{p}_i) - \boldsymbol{\hat{\xi}}_{ij}] .
\end{equation}
Here, $\boldsymbol{P}_{ij}$ is the covariance matrix corresponding to the noise of the relative measurements $\boldsymbol{\hat{\xi}}_{ij}$.

\subsection{Loop Closure Detection}

\begin{figure}[tb]
\centering
\includegraphics[width=0.445\textwidth]{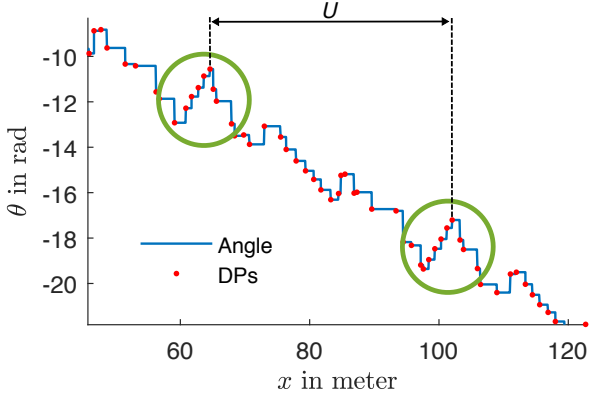}
\caption{Example for the piecewise linear orientation function $\theta(x)$. The green circled regions show similar path segments. The vertices or dominant points (DPs) of the pose graph are pictured as red dots. The estimated circumference $U$ for the closed environment is exemplarily depicted for a possible loop closing pair.}
\label{fig:PiecewiseOrientationFunction}
\end{figure}

\begin{figure}[tb]
\includegraphics[width=0.49\textwidth]{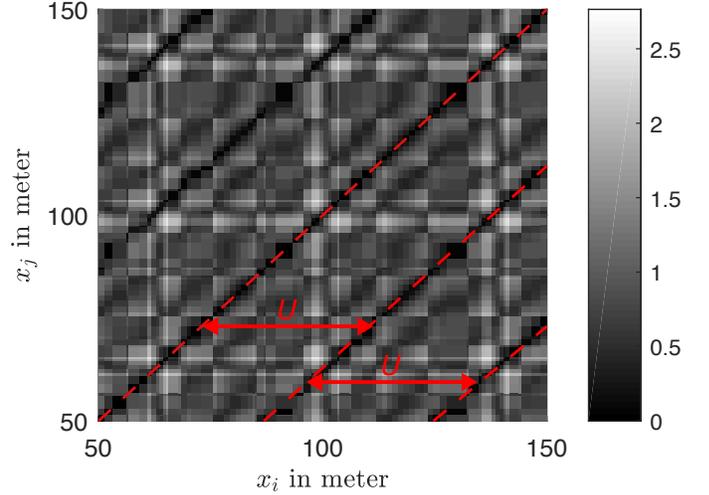}
\caption{Comparison error of the shapes of the neighborhood between the vertices of the pose graph. For better reading we plotted the error in the form $\log(1 - \boldsymbol{C}_{ij})$ and only a section of the matrix. The variables $x_i$ and $x_j$ are representing the position $l$ of the vertices $i$ and $j$ in meter along the pose graph. The estimated circumference $U$ for the closed environment can be read directly from the graphic.}
\label{fig:ComparisonError}
\end{figure}

\noindent Based on the pose graph, loop closing constraints are detected by comparing the shape of the neighborhood regions of each vertex with another. Therefore, a piecewise linear function
\begin{equation}
\label{eq:PiecewiseLinearFunctionOrientation}
\theta(x) = \phi_i \quad  \text{for}  \quad l_{i-1} \leq x < l_{i}, \quad i=0,1,\dots,N.
\end{equation}
representing the shape of the pose graph is constructed by accumulating the orientation and distance differences between the poses
\begin{equation}
\label{eq:cumulatedOrientationPathLength}
\begin{split}
\phi_{i} &= \phi_{i-1} + \Delta \phi_i \\
l_{i} &= l_{i-1} + ||\boldsymbol{v}_i|| .
\end{split}
\end{equation}
Here, ${\boldsymbol{v}_i = \boldsymbol{x}_{i} - \boldsymbol{x}_{i-1}}$ and ${\Delta \phi_i = \varphi_{i} - \varphi_{i-1}}$ starting by ${\phi_0 = \varphi_0}$ and ${l_0 = 0}$. \figref{fig:PiecewiseOrientationFunction} shows such a constructed piecewise orientation function. By defining the neighborhood of a vertex $i$ as ${[l_i - L_{\text{NH}},l_i + L_{\text{NH}}]}$, a comparison error between two vertices $i$ and $j$ is given as
\begin{equation}
\boldsymbol{C}_{ij} = \int_{-L_{\text{NH}}}^{+L_{\text{NH}}} \left[ \theta(l_i + x) - \phi_i \right] - \left[ \theta(l_j + x) - \phi_j \right] \text{d}x .
\label{eq:ComparisonErrorIntegral}
\end{equation}
We rewrite \eqref{eq:ComparisonErrorIntegral} as a sum over $m$ linearly distributed evaluation points
\begin{equation}
\boldsymbol{C}_{ij} = \frac{1}{m} \sum_{k=1}^m \left[ \theta(l_i + x_k) - \phi_i \right] - \left[ \theta(l_j + x_k) - \phi_j \right]
\end{equation}
with $x_1 = -L_{\text{NH}}$, $x_m = +L_{\text{NH}}$. In \figref{fig:ComparisonError}, a resulting error matrix between all vertices is graphically illustrated. A loop closing pair ${SP_k = \{\boldsymbol{p}_i,\boldsymbol{p}_j\}}$ for $i \neq j$ is defined as a local minimum of $C_{ij}$ for which holds $C_{ij,\text{min}} < c_{\text{min}}$. A local minimum represents thereby the best possible loop closure in a certain region of the error matrix and the threshold $c_{\text{min}}$ ensures that not every local minimum is selected as loop closing pair, but only sufficient accurate ones. Thus, the parameters $L_{\text{NH}}$ and $c_{\text{min}}$ are crucial for efficiently finding convenient loop closing pairs and will be learned through Bayesian Optimization. This process is discussed in \seref{subse:LCParams}.\\

\noindent After detecting a loop closure between the vertices $i$ and $j$ of the pose graph, the loop closing constraint as a relative measurement $\hat{\xi}_{ij}$ has to be added. Therefore, the neighborhood regions of both poses $i$ and $j$ are discretized as distinct points, represented by the sets $X_i = \{\boldsymbol{x}_{i,1},\dots,\boldsymbol{x}_{i,K}\}$ and $X_j = \{\boldsymbol{x}_{j,1},\dots,\boldsymbol{x}_{j,K}\}$, and transformed such that both poses $i$ and $j$ are equal with $\hat{\boldsymbol{p}}_{i} = \hat{\boldsymbol{p}}_{j} = [0,0,0]^{\top}$. By using an adapted ICP approach \cite{bergstrom2014robust}, which minimizes the distance error
\begin{equation}
\min_{\boldsymbol{R}_{\beta},\boldsymbol{t}} E_{\text{dist}}(\boldsymbol{R}_{\beta},\boldsymbol{t}) = \min_{\boldsymbol{R}_{\beta},\boldsymbol{t}} \sum_{k=1}^{K} ||\boldsymbol{R}_{\beta} \boldsymbol{x}_{i,k} + \boldsymbol{t} - \boldsymbol{x}_{i,k}^*||
\end{equation}
with $\boldsymbol{x}_{i,k}^*$ being the point of $X_j$ closest to $\boldsymbol{x}_{i,k}$, a two dimensional rotation $\boldsymbol{R}_{\beta}$ with $\beta$ being the rotation angle and a translation vector $\boldsymbol{t} = [t_x,t_y]^{\top}$ can be calculated. The loop closing constraint can then be derived using \eqref{eq:RelativeMeasurement} by transforming $\hat{\boldsymbol{p}}_{j}$ given the rotation and translation which leads to
\begin{equation}
\hat{\boldsymbol{\xi}}_{ij} = \begin{bmatrix}
t_x & t_y & \beta
\end{bmatrix}^{\top} .
\end{equation}
The corresponding covariance matrix can be calculated using the correlation error $C_{ij}$ and tuneable parameters $\gamma_1$ and $\gamma_2$
\begin{equation}
\label{eq:LoopClosingCovariance}
\boldsymbol{P}_{\text{lc},ij} = \text{diag}\left(
\begin{bmatrix}
\gamma_1 & \gamma_1 & \gamma_2
\end{bmatrix} \right)
C_{ij} .
\end{equation}
The parameters $\gamma_1$ and $\gamma_2$ are constant for all loop closing constraints and will be learned through Bayesian Optimization. This process is discussed in \seref{subse:PGOParams}. For the odometric constraints we generate the covariance matrices as
\begin{equation}
\label{eq:OdometricCovariance}
\boldsymbol{P}_{\text{odometric},ij} = \text{diag}\left(
\begin{bmatrix}
\cos(\varphi_i) (\alpha_3 \delta_T + \alpha_4 \delta_R) \\
\sin(\varphi_i) (\alpha_3 \delta_T + \alpha_4 \delta_R) \\
\alpha_1 \delta_R + \alpha_2 \delta_T
\end{bmatrix} \right)
\end{equation}
on the basis of the odometry model presented in \cite{thrun2002probabilistic} and under the assumption that only one translation $\delta_T$ and one rotation $\delta_R$ occur. The parameters $\alpha_1,\dots,\alpha_4$ can be learned. Here, we assume these parameters are given due to a known odometry model of the underlying differential drive system.

\subsection{Recurrent Symmetric Structures}\label{se:Recurrent}

\noindent A problem for the approach introduced above are recurrent symmetric structures. Such structures are present in many real world scenarios, and hence an autonomous robot needs to be able to cope with them. Therefore, we introduce a feasibility check
\begin{equation}
\label{eq:FeasibilityRecurrent}
| \pi - \text{mod}\left(\Delta \varphi_{ij}, 2 \pi \right) | > \varphi_{\text{cycle}}
\end{equation}
for every loop closing pair ${\{\boldsymbol{p}_i,\boldsymbol{p}_j\}}$ with the respective difference in orientation $\Delta \varphi_{ij} = \varphi_j - \varphi_i$. Here, $\text{mod}(a,b)$ is the modulo function which gives back the remainder of the Euclidean division of $a$ by $b$. Only loop closing pairs which pass the check of \eqref{eq:FeasibilityRecurrent} are considered for pose graph optimization. The feasibility check is based on the assumption, that, on average, the orientation error of the odometric measurements will sum up to zero. However, the orientation difference can largely differ and thus the meta-parameter $\varphi_{\text{cycle}} \in [0,\pi]$ has to be selected accordingly.

\section{Meta-Parameter Learning}\label{se:ParameterLearning}

\noindent To learn the unknown meta-parameters for the above mapping algorithm, we define an optimization problem with the objective
\begin{equation}
\label{eq:generalCostFunction}
\min_{\boldsymbol{\theta}} c(\boldsymbol{\theta})
\end{equation}
as a general cost function. This cost is then minimized through episodic BO \cite{shahriari2015taking} with expected improvement \cite{mockus1978application}. To optimize both terms, the loop closing parameters $L_{\text{NH}}$, $c_{\text{min}}$, $\varphi_{\text{cycle}}$ and the pose graph optimization parameters $\gamma_1$, $\gamma_2$ we define a two-stage optimization process. First, we optimize the loop closing parameters $L_{\text{NH}}$, $c_{\text{min}}$, $\varphi_{\text{cycle}}$ which gives us as a by-product an estimate of the circumference of the closed environment $U$. Based on the estimated circumference $U$ we can define a cost function for optimizing the pose graph parameters $\gamma_1$, $\gamma_2$. Hence, a joint optimization of all parameters is not suitable. In the following, we derive the two cost functions required for the optimization process. 

\subsection{Stage 1 -- Optimization of Loop Closing Parameters}\label{subse:LCParams}
\begin{figure}
\subfloat{\includegraphics[width=0.18\textwidth]{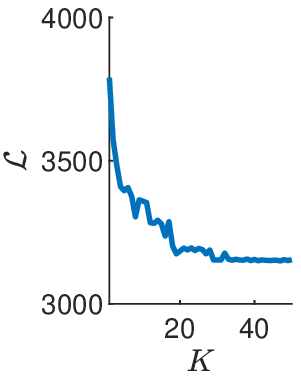}}\hfill
\subfloat{\includegraphics[width=0.30\textwidth]{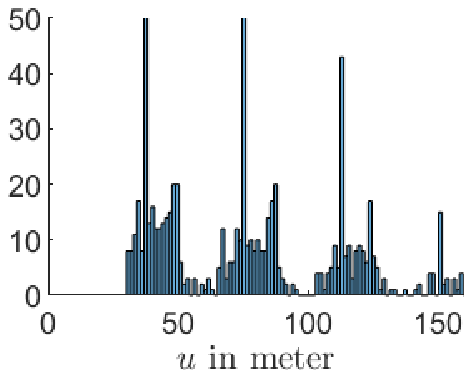}}\\
\subfloat{\includegraphics[width=0.18\textwidth]{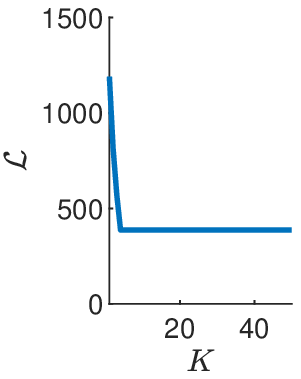}}\hfill
\subfloat{\includegraphics[width=0.30\textwidth]{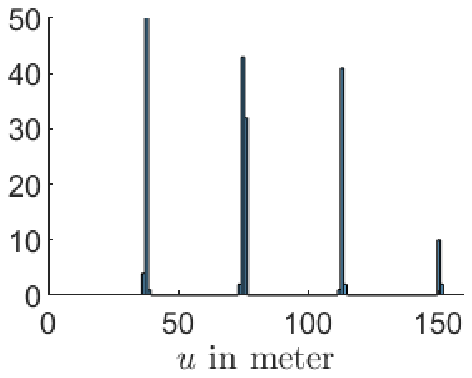}}
\caption{For the top panels, the mapping parameters have been ill-chosen. The upper left panel shows the negative log likelihood history and the upper right panel the histogram for the path distances of the loop closing pairs. In the bottom panels the mapping parameters have been well-chosen. Again, in the left panel the negative log likelihood history is shown and in the right the histogram.}
\label{fig:LikelihoodAndHistograms}
\end{figure}

\noindent We assume that the odometric error between two poses $i$ and $j$ is on average zero. This is a quite strong assumption, however, a non-zero mean value will be inherent in the generated map and thus compensated when navigating with the same robot odometry. To model this error, we use a Gaussian Distribution ${\boldsymbol{\epsilon}  \sim \mathcal{N}\left( \boldsymbol{0}, \boldsymbol{P}\right)}$ with the covariance matrix $\boldsymbol{P}$. Let $u$ then denote the distance along the pose graph between a loop closing pair $i$, $j$
\begin{equation}
\label{eq:DistanceBetweenLC}
u = \sum_{k=i}^{j-1} ||\boldsymbol{x}_{k+1} - \boldsymbol{x}_{k}|| .
\end{equation}
Given the assumption from above, the path distances for all loop closing pairs $\boldsymbol{u} = [u_1, u_2, \dots, u_M]$, identified by cycling around a closed environment, are, on average, multiples of the circumference $n U$. Here, $n \in \mathbb{N}^+$ is a positive integer, representing the number of cycles before the loop closure detection. Hence, if all loop closures are detected properly, a histogram of the path distances $\boldsymbol{u}$ has only equally distributed peaks at positions $nU$. The right panels of \figref{fig:LikelihoodAndHistograms} show such histograms for ill-detected loop closures (top) and well-detected loop closures (bottom). To transform this idea into a cost function, we can learn a Gaussian Mixture Model (GMM) \cite{reynolds2015gaussian} with the probability distribution
\begin{equation}
p(\boldsymbol{u}) = \sum_{k=1}^{K} \pi_k \mathcal{N}\left( \boldsymbol{u} | \mu_k, \Sigma_k \right)
\end{equation} 
from observed path distances $\boldsymbol{u}$. Here, $K$ is the number of mixture components and $\pi_k$, $\mu_k$, $\Sigma_k$ the mixture weight, the mean and the variance of the $k$-th component respectively. As part of the cost function we use the negative log likelihood of the GMM
\begin{equation}
\label{eq:LogLikelihoodGMM}
-\mathcal{L} = -\ln p(\boldsymbol{u}|\boldsymbol{\pi},\boldsymbol{\mu},\boldsymbol{\Sigma}) = -\sum_{i=1}^{M} \ln \left[ \sum_{k=1}^{K} \pi_k \mathcal{N}\left( u_i | \mu_k, \Sigma_k \right) \right]
\end{equation} 
over the data set ${\boldsymbol{u} = [u_1, u_2, \dots, u_M]}$. The log likelihood decreases if the dataset $\boldsymbol{u}$ meets the above assumption of evenly distributed peaks at positions $nU$. A common strategy for training GMMs is to iteratively increasing $K$ until the log likelihood does not improve further. In the left column in  \figref{fig:LikelihoodAndHistograms}, the evolution of the negative log likelihood with respect to the number of components of the GMM is shown. For fitting the GMM the iterative Expectation-Maximization (EM) algorithm is used \cite{dempster1977maximum}, \cite{mclachlan1988mixture}. The EM algorithm starts with a randomly selected model and then alternately optimizes the allocation of the data $\boldsymbol{u}$, i.e. the weighting $\pi_k$, to the individual parts of the model and the parameters of the model $\mu_k$ and $\Sigma_k$. If there is no significant improvement, the procedure is terminated. \\

\noindent We define the cost function for the loop closure detection as
\begin{equation}
\label{eq:CostLoopClosureDetection}
\min_{\boldsymbol{\theta}} c(\boldsymbol{\theta}) =  \min_{\boldsymbol{\theta}} \left(- \mathcal{L} - \log(M) \right),
\end{equation}
with the unknown parameters ${\boldsymbol{\theta} = [L_{\text{NH}},c_{\text{min}}]}$, the length of the neighborhood and the minimum comparison error, and $M$ being the number of loop closures found. The cost function represents a trade-off between the number of loop closures, where more reliable loop closures result in a better pose graph optimization, and a restrictive choice of loop closures to avoid false detection. Based on the best GMM fit, the circumference of the closed environment $U$ can be estimated.\\

\noindent To also learn the meta-parameter $\varphi_{\text{cycle}}$ for the feasibility check for recurrent symmetric structures, we can calculate the negative log likelihood for orientation differences of the loop closing pairs $\Delta \varphi_{ij}$, similar to \eqref{eq:LogLikelihoodGMM}, under the assumption that all $\Delta \varphi_{ij}$ for accurate loop closures are close to $n 2 \pi$ with $n \in \mathbb{N}^+$. \eqref{eq:CostLoopClosureDetection} than turns into
\begin{equation}
\min_{\boldsymbol{\theta}} c(\boldsymbol{\theta}) =  \min_{\boldsymbol{\theta}} \left( -\mathcal{L} - \mathcal{L}_{\varphi} - \log(M) \right),
\end{equation}
with ${\boldsymbol{\theta}}$ being now ${\boldsymbol{\theta} = [L_{\text{NH}},c_{\text{min}},\varphi_{\text{cycle}}]}$. The additional cost term $- \mathcal{L}_{\varphi}$ represents the negative log likelihood of the GMM from \eqref{eq:LogLikelihoodGMM} over the data set ${\Delta \boldsymbol{\varphi} = \{\Delta \varphi_1, \Delta \varphi_2, \dots, \Delta \varphi_M\}}$, thus
\begin{equation}
\label{eq:LogLikelihoodGMMPhi}
-\mathcal{L}_{\varphi} = -\ln p(\Delta \boldsymbol{\varphi}|\boldsymbol{\pi},\boldsymbol{\mu},\boldsymbol{\Sigma}).
\end{equation} 

\subsection{Stage 2 -- Optimization of Pose Graph Parameters}\label{subse:PGOParams}

\noindent Based on our assumption of a zero mean odometric error we can assume the estimated circumference $U$ from the first stage of our optimization process to be the true circumference of the closed environment. Hence, we can define a cost function for learning the pose graph optimization parameters ${\boldsymbol{\gamma} = [\gamma_1, \gamma_2]}$ as
\begin{equation}
\label{eq:optimizeForU}
\min_{\boldsymbol{\gamma}} c(\boldsymbol{\gamma}) = \min_{\boldsymbol{\gamma}} |U - \hat{U}|
\end{equation}
where $\hat{U}$ represents the estimated circumference after pose graph optimization. Thus, we punish deviations between the estimated circumference based on the original pose graph and the optimized one. In order to estimate the circumference after pose graph optimization, a fit onto GMMs is performed as proposed in \seref{subse:LCParams}.

%In addition, the odometry model parameters ${\boldsymbol{\alpha} = [\alpha_1,\dots,\alpha_4]}$, if unknown, can be learned in addition, which changes \eqref{eq:optimizeForU} to
%\begin{equation}
%\label{eq:optimizeForUWIthAlphas}
%\min_{\boldsymbol{\gamma},\boldsymbol{\alpha}} c(\boldsymbol{\gamma},\boldsymbol{\alpha}) = \min_{\boldsymbol{\gamma},\boldsymbol{\alpha}} |U - \hat{U}| .
%\end{equation}

\section{Results}\label{se:Results}

\noindent We evaluated the accuracy of the pose graph optimization (performance) and the generality of our approach in different environments (robustness). As a measure for the performance, we used an error metric based on the relative displacement between poses 
\begin{equation}
E_{\text{rel}}(\boldsymbol{\xi}) = \frac{1}{N} \sum_{i,j} \text{trans}\left( \boldsymbol{\xi}_{i,j} \circleddash \boldsymbol{\xi}_{i,j}^* \right)^2 + \text{rot}\left( \boldsymbol{\xi}_{i,j} \circleddash \boldsymbol{\xi}_{i,j}^* \right)^2
\end{equation}
as introduced in \cite{burgard2009comparison}. Here, $\boldsymbol{\xi}_{i,j}$ are the relative transformations after pose graph optimization, $\boldsymbol{\xi}_{i,j}^*$ ideally the true relative transformations and $\textit{trans}$ and $\textit{rot}$ separate  the translational and rotational components. Additionally, we used a second error metric for comparing results obtained on real lawns where the true poses of the robot are unknown but a groundtruth of the environment is available. Therefore, we constructed a polygon defined by the points $X$ out of the optimized pose graph data and compare this polygon with a polygon representing the groundtruth, $X_{\text{true}}$. We then transform
\begin{equation}
\label{eq:Transform}
X \leftarrow \boldsymbol{R} \cdot X + \boldsymbol{t},
\end{equation}
such that the deviation between the enclosed areas $A$, $A_{\text{true}}$ of the polygons 
\begin{equation}
\label{eq:DeltaA}
\Delta A = 1 - \frac{A_{\text{true}} \cap A_{\text{estimate}}}{A_{\text{true}} \cup A_{\text{estimate}}}
\end{equation}
is minimized. Here, $\boldsymbol{R}$ is a rotation matrix and $\boldsymbol{t}$ a translational vector. The minimized difference then serves as secondary error metric.\\

\noindent We compared to the original approach from \cite{rottmann2019loop} using hand crafted and learned parameters. The handcrafted parameters have been selected according to the following rules:\\
The neighborhood $L_{\text{NH}}$ should be chosen such that $2 L_{\text{NH}}$ is slightly larger then half of the true circumference. Thus, we like to use slightly more than $50 \, \%$ of $U$ for shape comparison. The comparison error threshold $c_{\text{min}}$ should be chosen according to the complexity of the given map.  A more complex map requires a larger comparison error threshold to account for more complicated comparisons. The meta-parameter $\varphi_{\text{cycle}}$ has to be chosen with regard to the recurrent symmetric structures of the given map. Here, only Map 1 has such structures with which we can cope by setting $\varphi_{\text{cycle}} = \pi/2$. The pose graph optimization parameters $\gamma_1, \gamma_2$ are kept constant with $\gamma_1 = 1$ and $\gamma_2 = 1$.

\subsection{Simulation}

\begin{figure}[tb]
\subfloat[Map 1]{\includegraphics[width=0.17\textwidth]{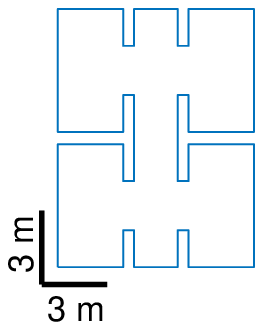}}
\subfloat[Map 2]{\includegraphics[width=0.17\textwidth]{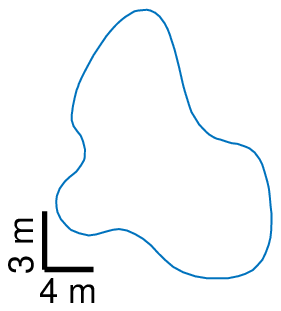}}
\subfloat[Map 3]{\includegraphics[width=0.17\textwidth]{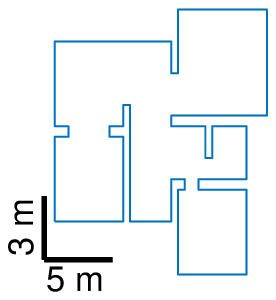}}
\caption{Simulation environments used for evaluating the proposed learning procedure for mapping in closed environments. From left to right: A symmetric environment ($U = 77 \, m$), a curved environment ($U = 52 \, m$) and an apartment environment ($U = 100 \, m$).}
\label{fig:ExampleMaps}
\end{figure}

\noindent We show the robustness of the approach applying our mapping procedure in different simulated closed environments with hard features, such as recurrent structures, large dimensions or curvatures. For the simulation environment, we used the odometry motion model presented in \cite{thrun2002probabilistic}. We calibrated the odometry model by tracking lawn mower movements using a visual tracking system (OptiTrack) and computed the parameters using maximum likelihood estimation \cite{myung2003tutorial}. The calibrated parameters for the Viking MI 422P robot are presented in \tabref{tab:ParamModels} and are used for the simulation. To generate movement data, we used a wall-following algorithm cycling for $T = 2000\,s$ along the boundary of the closed environment. We statistically evaluated our approach simulating $20$ runs with a maximum of 30 iterations for the Bayes Optimizer. This optimization is designed for global optimization of black-box functions and does not require any derivatives.\\

\begin{table}[tb]
\centering
\caption{Measured Parameters for the odometry motion model \cite{thrun2002probabilistic}.} 
\label{tab:ParamModels}
\begin{tabular}{c|c|c|c}
$\alpha_1$ & $\alpha_2$ & $\alpha_3$ &  $\alpha_4$  \\ 
\hline 
0.0849 & 0.0412 & 0.0316  & 0.0173 \\ 
\end{tabular}
\end{table}

\noindent In \tabref{tab:simulationResults}, the simulation results for the different maps from \figref{fig:ExampleMaps} with different combinations of hand-crafted parameters are presented. Our adjusted approach, using the ICP method, clearly outperforms the original method. In all $20$ runs it leads to better map estimates after pose graph optimization. Moreover, learning the parameters enables the algorithm to generalize to different environments without prior knowledge about the odometry error, the shape or the circumference of the environment. This prior knowledge is essential for choosing suitable hand-crafted parameters. Without such knowledge, the parameters have to be manually tuned which might lead to disastrous mapping results. For example, changing the neighborhood parameter $L_{\text{NH}}$ for Map 3 to $L_{\text{NH}} = 15$ results in a large increase of the mapping errors. In addition, a change in odometry error accuracy can be compensated by learning the mapping meta-parameters, as demonstrated in simulations with odometry model parameters $\alpha_i = 0.1,0.2$ for $i=1,\dots,4$.

\begin{figure}[tb]
\subfloat[Odometry Measurements]{\includegraphics[width=0.23\textwidth]{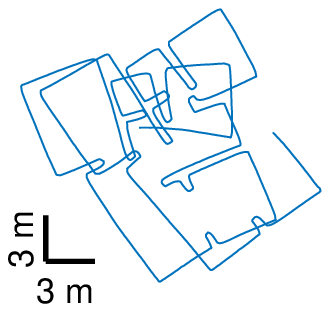}}
\subfloat[Original Approach]{\includegraphics[width=0.23\textwidth]{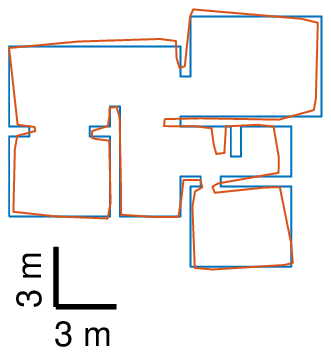}}\\
\subfloat[Adjusted Approach]{\includegraphics[width=0.23\textwidth]{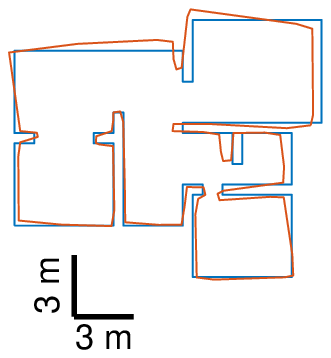}}
\subfloat[Learned Parameters]{\includegraphics[width=0.23\textwidth]{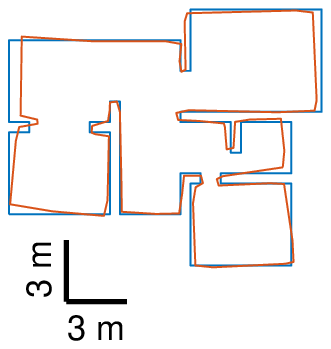}}
\caption{Exemplary mapping results with the simulation environment ''Map 3'' and an odometry error of $\alpha_i = 0.2$. In (b)-(d) the blue line shows the true shape of the environment and the red line the map estimate.}
\label{fig:ExampleMappingResults}
\end{figure}

%\begin{table}[t]
%\centering
%\caption{Performance of the mapping approach for different odometry erros with map 3. A fail means that no valid map could have been generated.}
%\label{tab:DiffOdometryErrors}
%\begin{tabular}{l|l|l}
%$\alpha_i$ & Hand crafted parameters \cite{rottmann2019loop} & BO approach\\
%\hline
%0.1 & $8.14 \% \pm 2.24 \%$ & $\boldsymbol{7.58} \% \pm 2.22 \%$\\
%0.2 & $12.74 \% \pm 2.70 \%$ + 1 fail & $\boldsymbol{11.33} \% \pm 2.34 \%$ \\
%0.3 & $19.20 \% \pm 4.42 \%$ & $\boldsymbol{16.76} \% \pm 4.51 \%$\\
%0.4 & $18.05 \% \pm 4.85 \%$ + 1 fail & $\boldsymbol{19.50} \% \pm 5.43 \%$ + 1 fail\\
%0.5 & $21.59 \% \pm 5.64 \%$ + 7 fails & $\boldsymbol{24.05} \% \pm 8.24 \%$\\
%\end{tabular}
%\end{table}

\begin{table*}[t]
\centering
\caption{Simulation results for different maps and hand-crafted parameters for the original approach from \cite{rottmann2019loop}, the adapted approach and with learned parameters. The table shows the mean and the standard deviations for the relative displacement errors.\\
$^*$the measured odometry parameters from \tabref{tab:ParamModels} are used}
\label{tab:simulationResults}
\begin{tabular}{c|c|c|c|c|c|c|c|c|c}
\multirow{2}{*}{Map} & \multirow{2}{*}{$L_{\text{NH}}$} & \multirow{2}{*}{$c_{\text{min}}$} & \multirow{2}{*}{$\alpha_i$} & \multicolumn{2}{c|}{Original Approach} & \multicolumn{2}{c|}{Adjusted Approach} & \multicolumn{2}{c}{Learned Parameters} \\
 & & & & $E_{\text{trans}}$ & $E_{\text{rot}}$ & $E_{\text{trans}}$ & $E_{\text{rot}}$ & $E_{\text{trans}}$ & $E_{\text{rot}}$ \\
\hline
\hline
 1 & 20 & 1.0 & $^*$ & $0.0036 \pm 0.0028$ & $0.0076 \pm 0.0053$ & $\bm{0.0021} \pm 0.0034$ & $\bm{0.0052} \pm 0.0055$ & $0.0006 \pm 0.0015$ & $0.0093 \pm 0.0166$ \\
 2 & 15 & 0.5 & $^*$ & $0.0406 \pm 0.0943$ & $0.0569 \pm 0.0399$ & $0.0183 \pm 0.0068$ & $0.0564 \pm 0.0404$ & $\bm{0.0002} \pm 0.0005$ & $\bm{0.0003} \pm 0.0003$ \\
 3 & 30 & 0.3 & $^*$ & $0.1290 \pm 0.5532$ & $0.0049 \pm 0.0098$ & $\bm{0.0020} \pm 0.0050$ & $\bm{0.0021} \pm 0.0021$ & $0.0024 \pm 0.0036$ & $0.0262 \pm 0.0780$ \\
  \hline
  3 & 30 & 1.5 & $^*$ & $2.787 \pm 12.16$ & $0.0095 \pm 0.0093$ & $0.5442 \pm 2.004$ & $0.0063 \pm 0.0136$ & $\bm{0.0017} \pm 0.0026$ & $\bm{0.0335} \pm 0.0607$ \\
    3 & 15 & 0.3 & $^*$ & $35.57 \pm 158.0$ & $0.0191 \pm 0.0214$ & $31.03 \pm 137.6$ & $0.0139 \pm 0.0220$ & -- & -- \\
 \hline
 3 & 30 & 0.3 & 0.1 & $0.0151 \pm 0.0418$ & $0.0060 \pm 0.0086$ & $0.0085 \pm 0.0296$ & $0.0025 \pm 0.0020$ & $\bm{0.0070} \pm 0.0079$ & $\bm{0.0665} \pm 0.1932$ \\
 3 & 30 & 0.3 & 0.2 & $44.54 \pm 188.8$ & $0.0304 \pm 0.0580$ & $1.65 \pm 6.88$ & $0.0158 \pm 0.0285$ & $\bm{0.0205} \pm 0.0387$ & $\bm{0.0352} \pm 0.0642$ \\
\end{tabular}
\end{table*}

%\noindent We finally evaluated the influence of the different parameter tuni
%
%\begin{table}[h!]
%\centering
%\caption{Performance of the mapping approach for different combinations of parameter optimization with map 3 and standard odometry error.}
%\label{tab:SimResultsDiffParamOpt}
%\begin{tabular}{l|l}
%Optimized Parameter & Mapping Error \\
%\hline
%$l_{\text{NH}}$, $c_{\text{min}}$, $\gamma_1$, $\gamma_2$ & $5.93 \% \pm 1.66 \%$ \\
%$l_{\text{NH}}$, $c_{\text{min}}$ & $0.00 \% \pm 0.00 \%$\\
%$\gamma_1$, $\gamma_2$ & $0.00 \% \pm 0.00 \%$\\
%\end{tabular}
%\end{table}

\subsection{Real Data}

\noindent For generating real data, we drove the lawn mower along the boundary line of two different lawn areas. The velocity of the lawn mower driving along the boundary has been set to $\SI{0.3}{\metre\per\second}$. The odometry data has been sampled with a frequency of approximately $\SI{20}{Hz}$. 

\begin{figure*}
\subfloat[The courtyard of our Institute. We used the inner lawn area for testing the proposed mapping method.]{\includegraphics[width=0.32\textwidth]{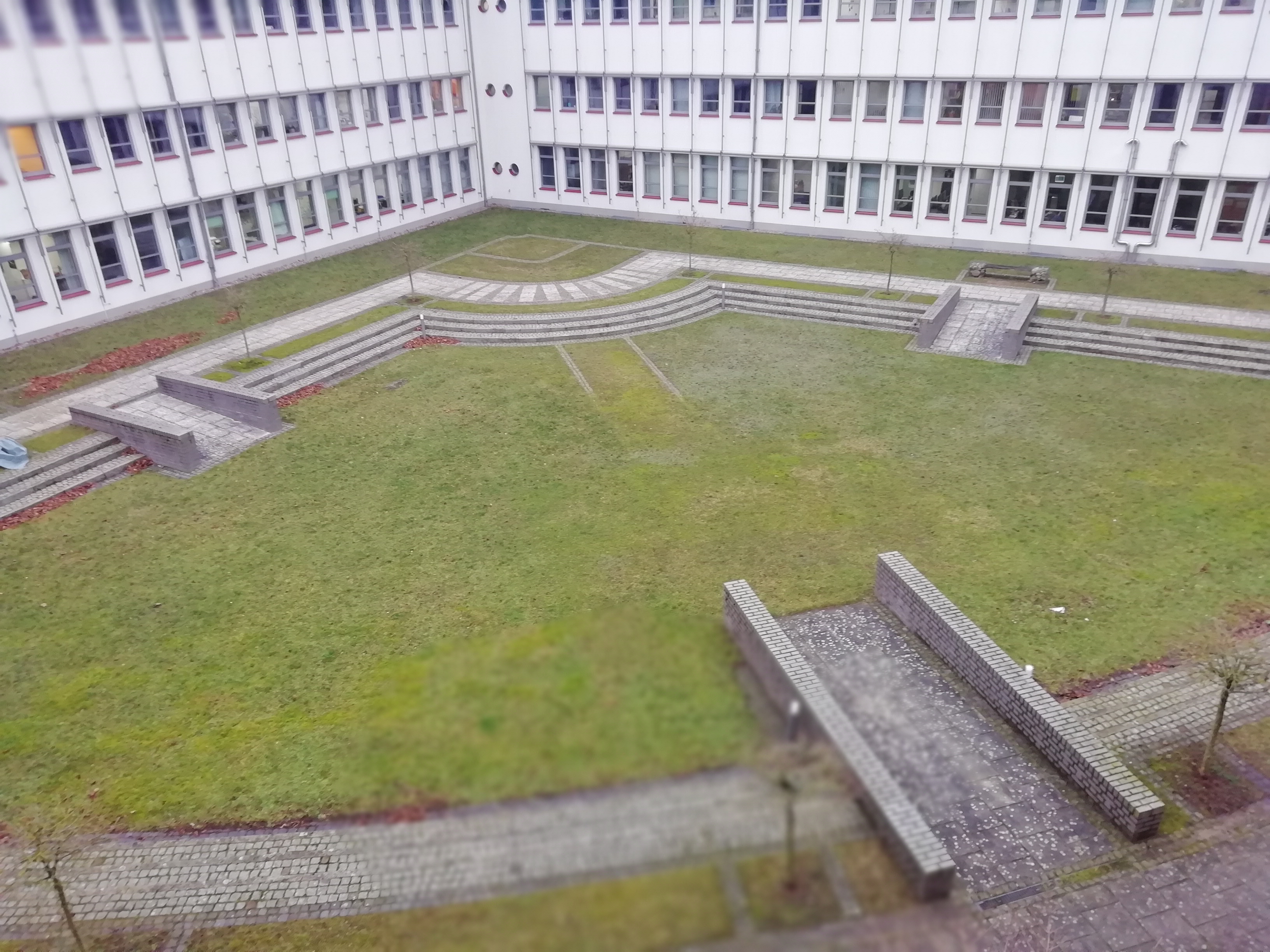}}
\hspace{0.01\textwidth}
\subfloat[The estimated path of the robot generated from its wheel odometry.]{\includegraphics[width=0.32\textwidth]{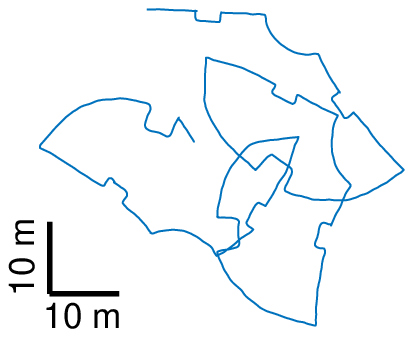}}
\hspace{0.01\textwidth}
\subfloat[The estimated map (red) and the true shape of the test environment (blue).]{\includegraphics[width=0.32\textwidth]{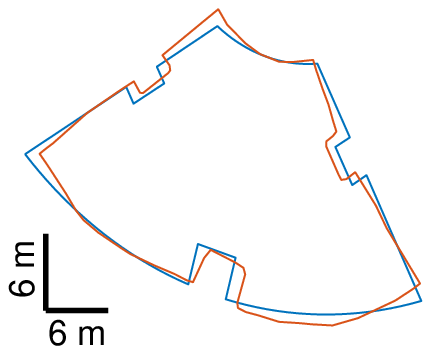}}
\caption{The real courtyard depicted (a), the collected odometry data (b) and the map estimate with learned parameters (c).}
\label{fig:ExampleCourtyard}
\end{figure*}

\noindent In \figref{fig:ExampleCourtyard}, the university courtyard, the measured odometry data and the generated map estimate are shown. The ground truth is available as CAD data, such that we can compare our map estimations using \eqref{eq:DeltaA}. Based on the circumference $U = 106.8 \, m$ and the complexity of the environment, the hand crafted parameters have been set to $L_{\text{NH}} = 30$, $c_{\text{min}} = 0.3$. The resulting mapping error for the original approach is $\Delta A = 11.49 \%$, for the adjusted approach $\Delta A = 9.77 \%$ and the mapping error with learned parameters $c_{\text{min}} = 0.1967$, $L_{\text{NH}} = 32.32$, $\varphi_{\text{cycle}} = 1.57$, $\gamma_1 = 0.0104$, $\gamma_2 = 0.0122$ is $\Delta A = 9.24 \%$. Again, the adjusted approach outperforms the original approach and learning the parameters with the proposed cost functions leads to sufficiently accurate results.\\

\begin{figure*}
\subfloat[The top view onto a part of a lawn of a typical private household.]{\includegraphics[width=0.32\textwidth]{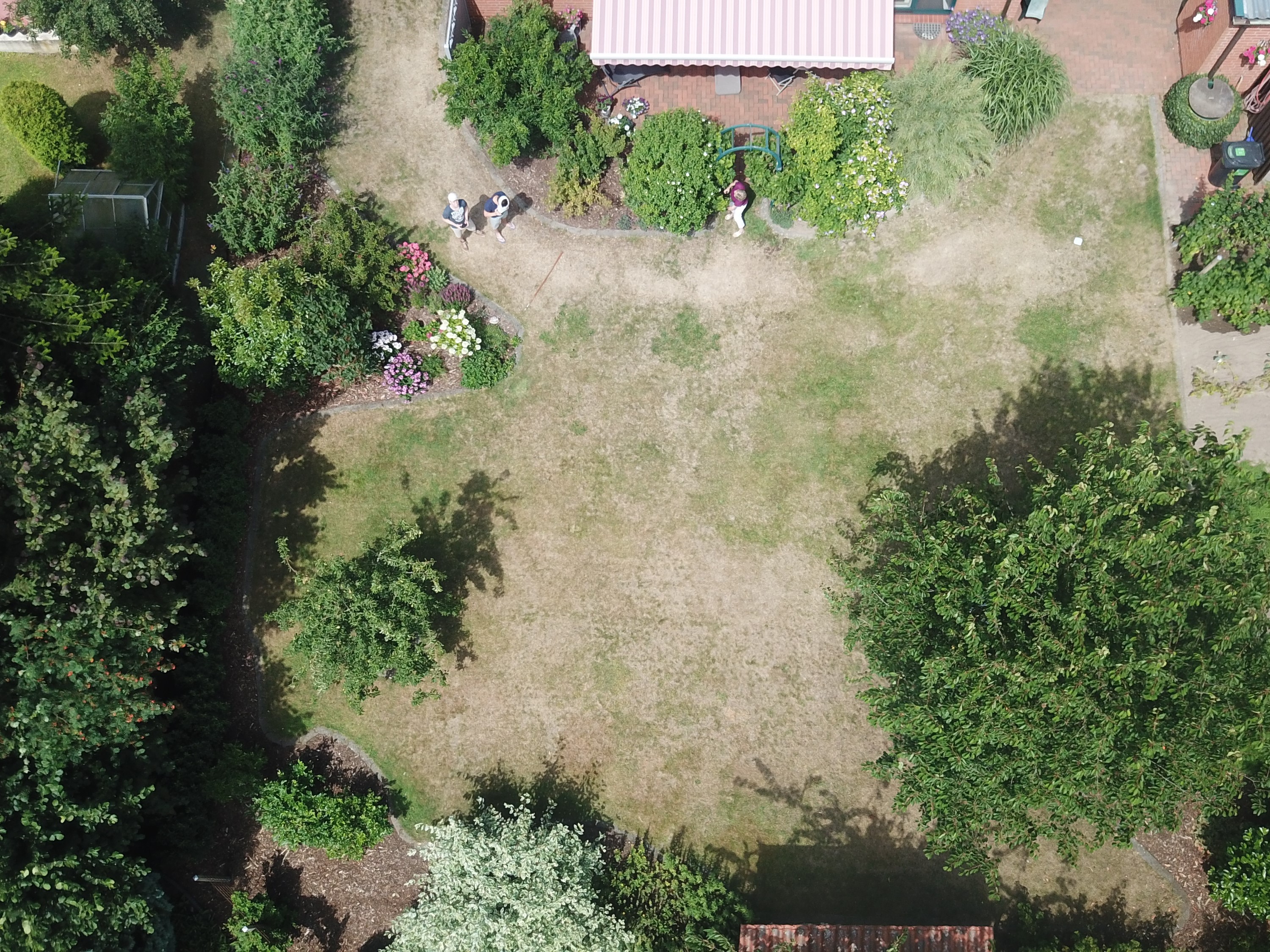}}
\hspace{0.01\textwidth}
\subfloat[The estimated path of the robot generated from its wheel odometry.]{\includegraphics[width=0.32\textwidth]{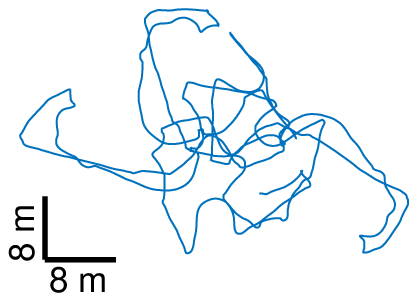}}
\hspace{0.01\textwidth}
\subfloat[The estimated map.]{\includegraphics[width=0.32\textwidth]{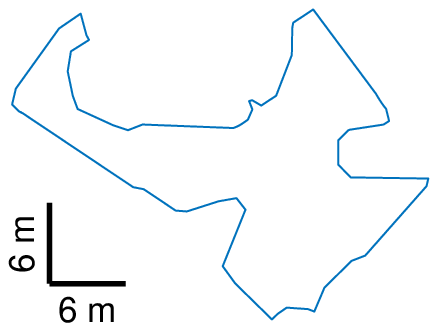}}
\caption{A typical lawn (a), the collected odometry data (b) and the map estimate with learned parameters (c).}
\label{fig:ExampleGarden}
\end{figure*}

\noindent In addition, we evaluated the mapping approach in a second real environment, a representative of a typical private lawn. In \figref{fig:ExampleGarden} from left to right, we show a part of the private lawn, the measured odometry data and the map estimate. Since we do not have ground truth data for this lawn, we compared the map results qualitatively with the image of the real garden. As demonstrated, the approach is capable of mapping large closed environments with narrow corridors based on severely distorted odometry data.

\section{Conclusion}\label{se:Conclusion}

\noindent Towards efficient localization and planning for low-cost robots, a first step is the generation of an accurate map estimate of the enclosed environment. Thereby, the robot has to learn required meta-parameters automatically to be able to adapt to different environments. Here, we have made improvements to the mapping algorithms for closed environment introduced in \cite{rottmann2019loop}, which significantly enhance the performance by allowing the algorithm to cope with recurrent symmetric structures as well as reducing the relative displacement error. Moreover, we proposed a cost function for meta-parameter learning for mapping algorithms in closed environments. This cost function does neither require any a-priori information about the environment nor domain expert knowledge and thus enables the robot to act truly autonomously. We demonstrated the feasibility, robustness and performance of our approach in both simulated and real closed environments. Thereby, we showed that based on the proposed mapping procedure, accurate map estimates of underlying closed environments can be produced. These map estimates are the first step towards intelligent behavior for low-cost robots, such as autonomous lawn mowers.\\

\subsection{Discussion}

\noindent The underlying assumption of a zero mean odometry error is quite strong and might not hold true under many circumstances, for example if one of the wheels is slightly smaller (e.g. due to air pressure). However, fusing the wheel odometry with IMU measurements, we are able to compensate for such inaccuracies. Moreover, we can detect wheel slippage. Otherwise, a non-zero odometric mean error will be inherited in the final map estimate and thus compensated by navigating with the same robot odometry.\\

\noindent In future work, we will investigate the possibilities of probabilistic approaches for efficiently mowing the lawn with high-confidence. Therefore, coverage grid maps with ''already mown lawn" probabilities similar as in \cite{hess2014probabilistic} can be used in combination with an adjusted intelligent complete coverage path planning algorithm, e.g. neural network approach \cite{yang2004neural}. Thereby, the ''mowing probabilities'' of the grid map are actualized based on a particle filter estimation.

\newpage

\balance

\bibliography{papers}
\bibliographystyle{plain}

\end{document}